\documentclass[letterpaper, 10 pt, conference]{ieeeconf}  
\usepackage{draftwatermark}
\SetWatermarkText{CONFIDENTIAL. Limited circulation. For reference of ICRA 2016 only}
\SetWatermarkLightness{0.5}
\SetWatermarkScale{0.15}

\IEEEoverridecommandlockouts                              

\newcommand{\tabincell}[2]{\begin{tabular}{@{}#1@{}}#2\end{tabular}}

\title{\LARGE \bf
Design, Modeling and Control of A Novel Amphibious Robot \\with Dual-swing-legs Propulsion Mechanism
}

\author{Yang Yi,  \emph{Member}, \emph{IEEE}, Zhou Geng, Zhang Jianqing, Cheng Siyuan, Fu Mengyin
\thanks{*~~The work was partly supported by National Natural Science Foundation of China (Grant No. NSFC 61473042 and 61105092) and Beijing Higher Education Young Elite Teacher Project (No.YETP1215). The
authors are with the School of Automation and State Key Laboratory
of Complex System Intelligent Control and Decision, Beijing Institute of Technology, Beijing 100081, China (phone: 86-10-68913985; fax: 86-10-68918381; e-mail: yang\_yi@bit.edu.cn).}
}
\usepackage{xcolor}
\usepackage{subfigure}
\usepackage{stfloats}
\usepackage{graphicx}
\graphicspath{{Pictures/}} 
\usepackage{makecell} 
\usepackage{multirow}
\usepackage{amsmath}
\usepackage{amssymb}
\begin{document}

\maketitle
\thispagestyle{empty}
\pagestyle{empty}

\begin{abstract}

This paper describes a novel amphibious robot, which adopts a dual-swing-legs propulsion mechanism, proposing a new locomotion mode. The robot is called FroBot, since its structure and locomotion are similar to frogs. Our inspiration comes from the frog scooter and  breaststroke. Based on its swing leg mechanism, an unusual universal wheel structure is used to generate propulsion on land, while a pair of flexible caudal fins functions like the foot flippers of a frog to generate similar propulsion underwater. On the basis of the prototype design and the dynamic model of the robot, some locomotion control simulations and experiments were conducted for the purpose of adjusting the parameters that affect the propulsion of the robot. Finally, a series of underwater experiments were performed to verify the design feasibility of FroBot and the rationality of the control algorithm.

\end{abstract}

\section{Introduction}

As a result of long process of evolution, some animals have possessed smart motion mechanisms and astute movement patterns. For instance, amphibians are able to move freely in the amphibious environment due to their unique structure and perception. Therefore, the design objective of amphibious robots is supposed to be finding appropriate movement patterns, optimizing mechanical structure and improving sensory ability. 

\begin{figure}[thpb]
      \centering
      \includegraphics[width=80mm]{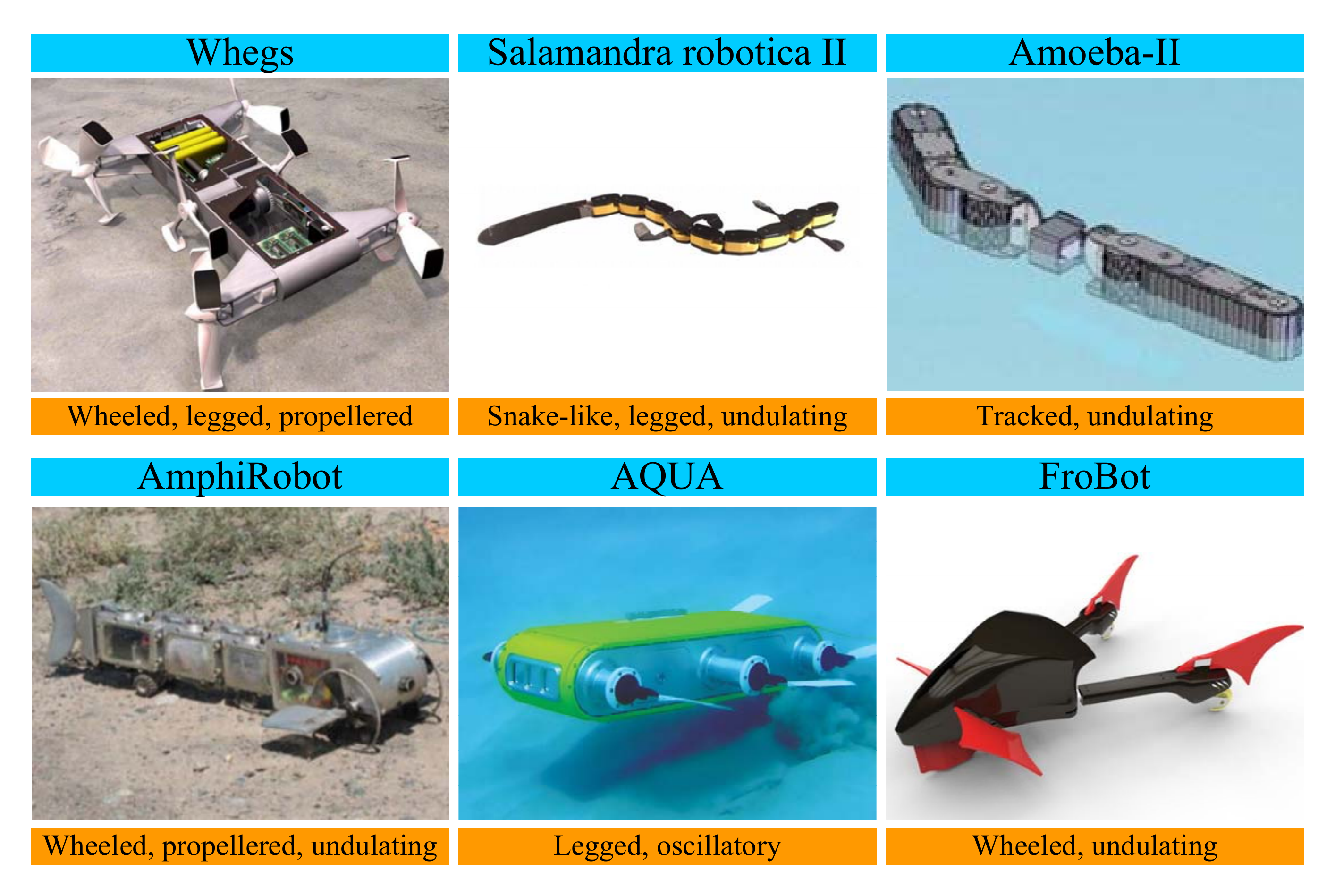}  
      \caption{Amphibious robots combine different locomotion modes}
      \label{Fig:FroBot}
\end{figure}

Locomotion modes of robots that move on land or in aquatic media mainly include wheeled, tracked, legged \cite{c1}, snake-like \cite{c2}, undulating \cite{c3}, oscillatory \cite{a1}, propellered, jet fluidic \cite{c4} mode and so on. Different locomotion modes have different advantages and disadvantages in different fields, which poses a big challenge in the design of an amphibious robot. So many amphibious robots adopt two or more locomotion modes to get the thrust in the amphibious environment, as shown in Fig. \ref{Fig:FroBot}. Whegs \cite{c5}, an amphibious robot that combines propellers with legs, can swim as well as walk. The special mechanism design of the robot combines the superior mobility of a leg with the controllability of a wheel. Salamandra robotica II \cite{c6}, another amphibious robot that makes a combination of three locomotion modes: legged, snake-like and undulating, can walk on rough road, maneuver   underwater and pass through transitional terrains such as sand and mud. An amphibious robot called Amoeba-II \cite{c7}, which combines tracked and undulating modes, can perform high maneuverability in amphibious environment.  In \cite{c8} is presented an amphibious robotic fish with wheel-propeller-fin mechanisms. This robot is  able to not only move on various terrains, but also perform fish-like and dolphin-like swimming. AQUA  \cite{c9}, a six-legged amphibious robot that stems from Rhex  \cite{c10,c11}, possesses remarkable amphibious locomotion performance. On account of the excellent design, AQUA can change the gait to adapt to different types of environments. In addition, there are also many outstanding amphibious robots, such as Omni-Paddle \cite{c12}, Groundbot \cite{c13}, Perambulator \cite{c14}, SeaDog \cite{c15}.

In this paper, we present a novel frog-inspired amphibious robot named FroBot, as shown in Fig. \ref{Fig:FroBot}. Different from most of other frog-inspired robots which can jump \cite{c16,c17}, we focus on the swimming movement of a frog and the breaststroke-like locomotion mode is innovatively applied to the amphibious robot. Based on the swing leg mechanism, this robot is designed to combine the advantages of wheeled locomotion and undulating locomotion. An unusual universal wheel structure is used to generate propulsion on land, while a pair of flexible caudal fins functions like the foot flippers of a frog to generate similar propulsion underwater. Wheeled vehicles have faster speed and higher carrying capacity than other locomotion modes on land, so they are quite suitable for moving on smooth surfaces. Among the numerous movement types, piscine undulating motion is always a research hotspot for zoologists, marine biologists and engineers. Undulating motion has great advantages over manmade propeller. It not only has higher energy conversion rate and the advantage of lighter noise, but also can use the energy of surrounding fluid. There are mainly two undulation types: Body and/or Caudal Fin (BCF) and Median and/or Paired Fin (MPF) \cite{c18}. As for FroBot, a pair of caudal fins is used to generate propulsion during the swinging of legs. In \cite{c19} is proposed a prototype of an underwater microrobot utilizing ICPF (Ionic Conducting Polymer Film) actuator as the servo actuator. There is a pair of fins with two ICPF actuators on it, and its swimming speed and orientation can be controlled by changing the frequency of input voltage of the ICPF actuator. This robot gives much inspiration to the design and control of FroBot.

Compared with the other amphibious robots, FroBot has the same dual-swing-legs propulsion mechanism in the amphibious environment and fewer actuators to move forward, making its control and manufacture easy to realize. It can move fast on smooth surfaces and have a high carrying capacity. The symmetrical dual-swing-legs locomotion mode can benefit the heading stability and it is an effective way to keep balance underwater. While due to the defect of the wheeled vehicles, FroBot has limited locomotion capacity in a complex terrain. Its structure, dynamic model and some related experiments will be presented as follows.

\section{Structure of The Amphibious Robot}

\begin{figure}[thpb]
      \centering
      \includegraphics[width=80mm]{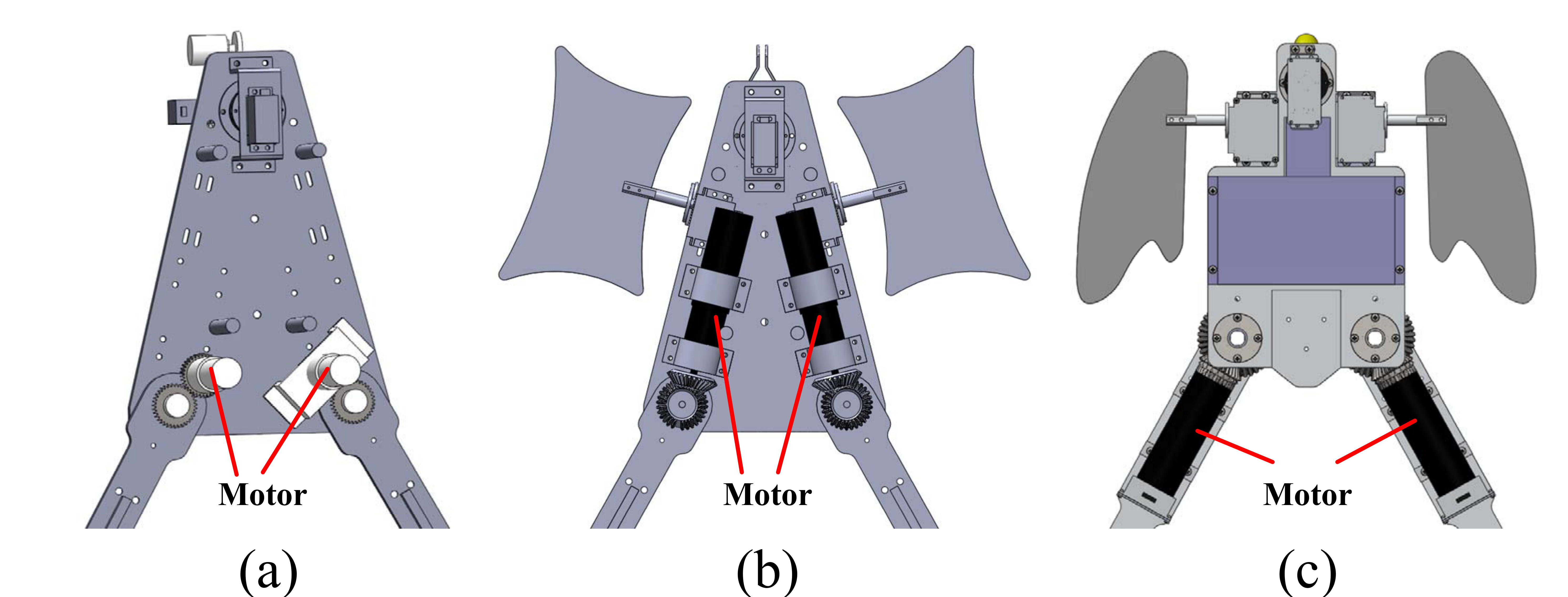}  
      \caption{Development of FroBot. (a) the first generation, (b) the second generation, (c) the third generation.}
      \label{Fig:The development}
   \end{figure}

From the first generation of land locomotion version to the second and the third generation of amphibious locomotion version, FroBot has been improved a lot. The first generation utilizes two motors mounted vertically to actuate the legs (see Fig. \ref{Fig:The development} (a)). For easy sealing, all motors are installed horizontally (see Fig. \ref{Fig:The development} (b)). For the purpose of better mass distribution, two driving motors can also be fixed on the legs (see Fig. \ref{Fig:The development} (c)). Unless otherwise stated, the analysis of underwater locomotion is based on the second generation.

\begin{figure}[thpb]
      \centering
      \includegraphics[width=80mm]{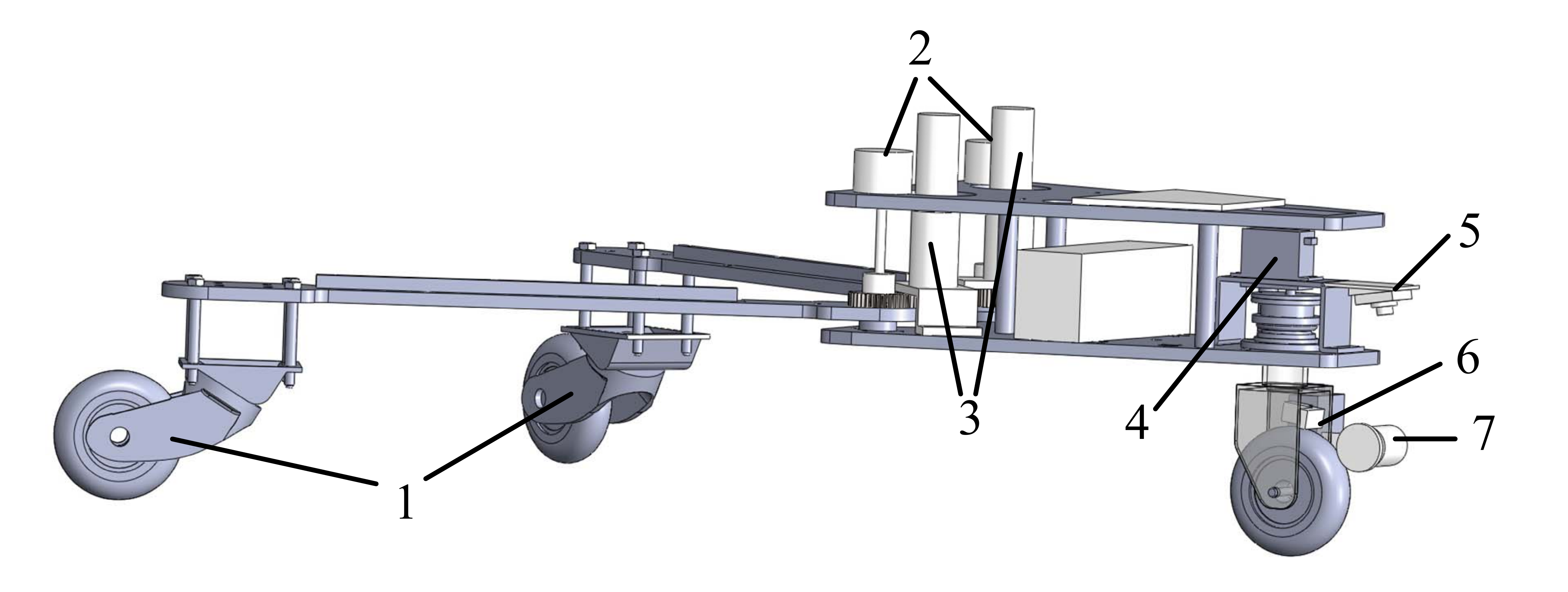}  
      \caption{The structure of the first generation. The robot mainly consists of two universal wheels (1), two none-contact hall angle sensors (WDH35) (2) to detect realtime swing angles, two DC motors (3), a steering-engine (4) for yaw movement, a linear array CCD (TSL1401) (5) used to detect trajectory position during the trajectory following experiment, a brake apparatus (6) controlled by a steering-engine, an encoder (7) used for velocity measurement. The microcontroller is MC9S12XS128, which transmits experiment data to computer using wireless bluetooth module.}
      \label{Fig:The structure}
   \end{figure}

The locomotion mode of FroBot on land is different from the traditional wheeled vehicles. It has a strong nonlinear character due to the installation pattern of its universal wheels. In order to get better control of the robot on land, we conducted several experiments based on the first generation of FroBot (see Fig. \ref{Fig:The structure}).

\begin{figure}[thpb]
      \centering
      \includegraphics[width=70mm]{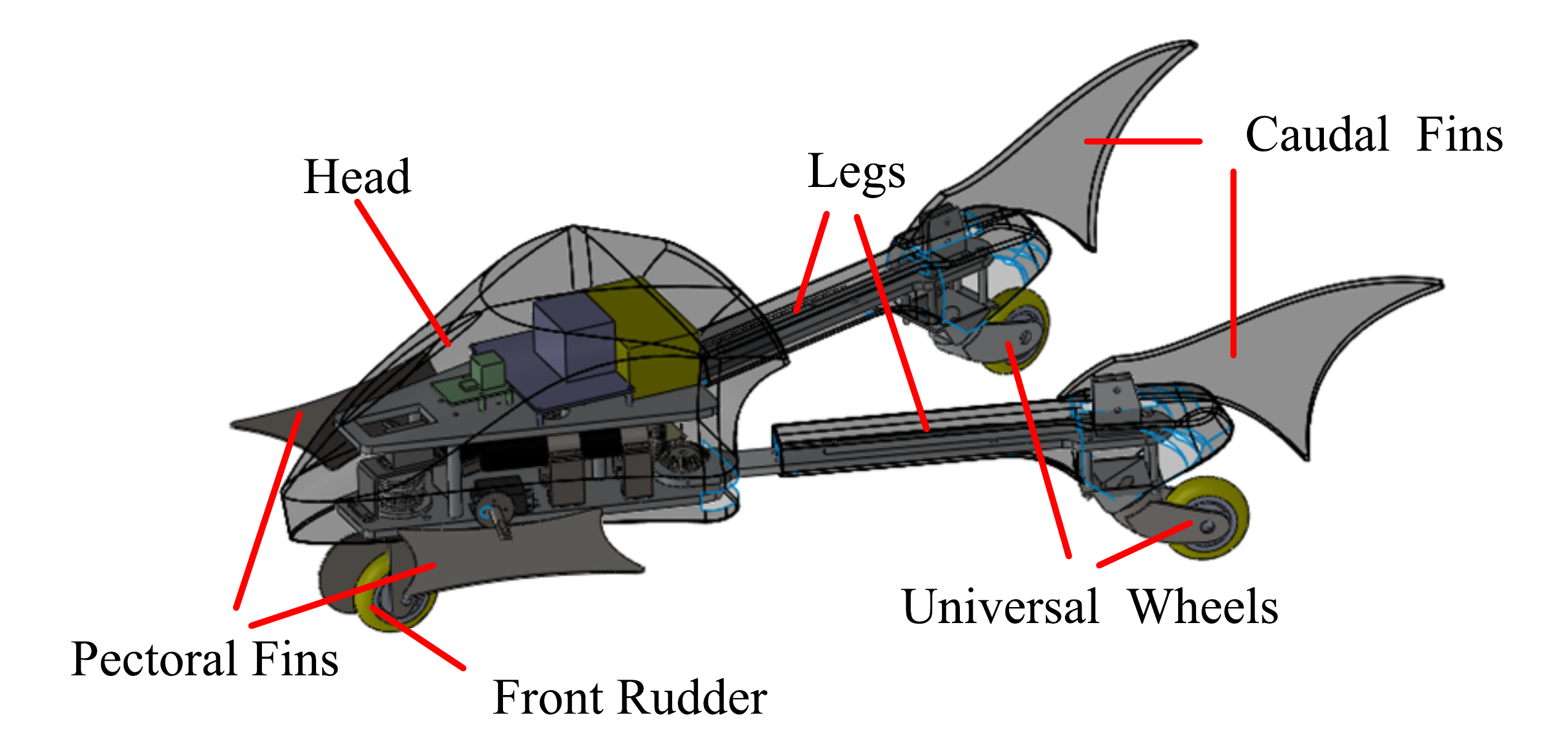}  
      \caption{Overall mechanical structure}
      \label{Fig:mechanical structure}
   \end{figure}

\begin{figure}[thpb]
      \centering
      \includegraphics[width=60mm]{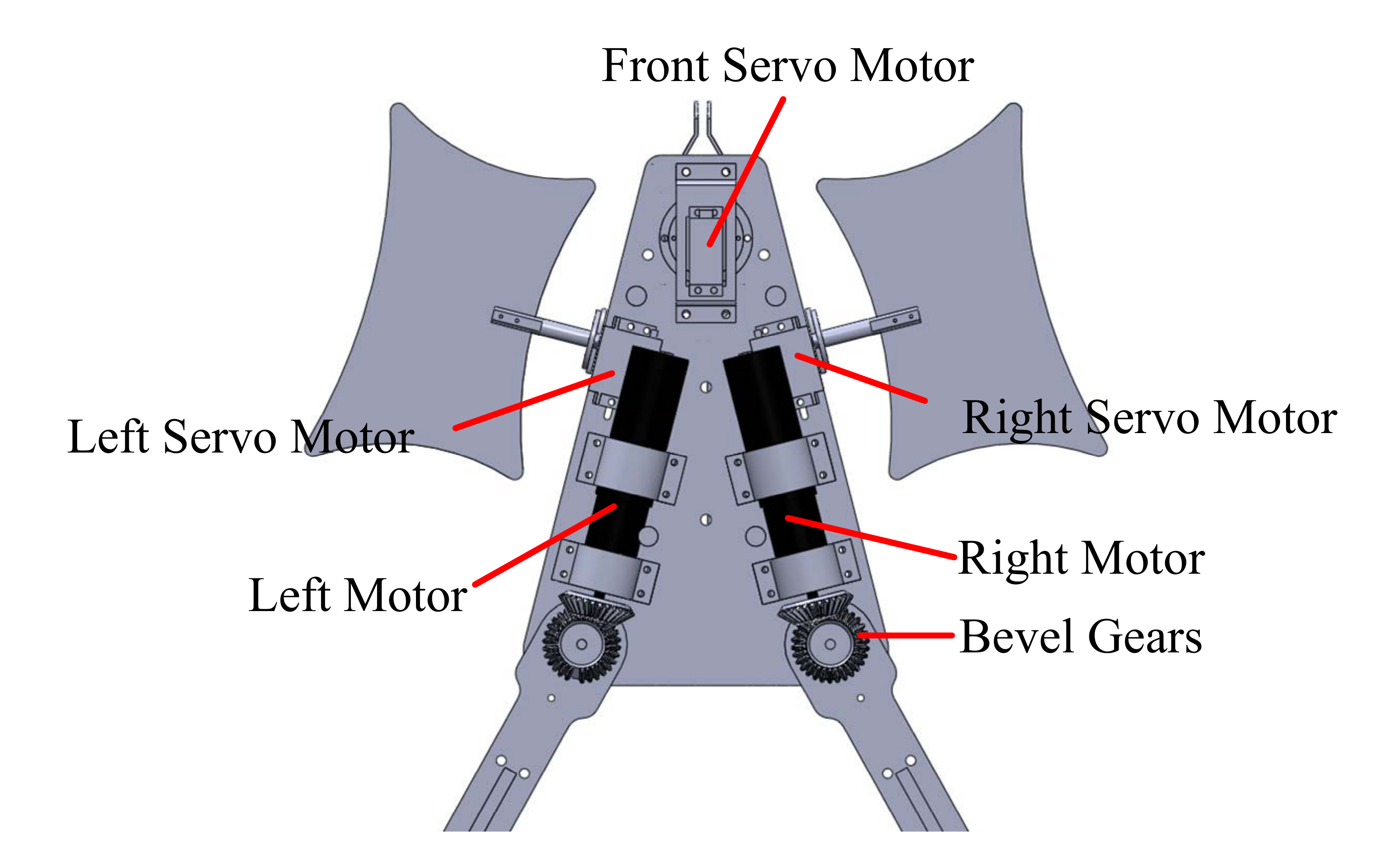}  
      \caption{Actuators of FroBot}
      \label{Fig:Actuators}
   \end{figure}

As shown in Fig. \ref{Fig:mechanical structure}, the robot is mainly composed of a head and a pair of legs. There is a front rudder with hybrid wheel-blade mechanism at the bottom of the head. The robot is equipped with one pectoral fin at each side of the head to achieve pitch and roll movement underwater. There is a pair of universal wheels and caudal fins at the end of legs. The Frobot's skeleton is made of aluminum alloy, and its outside is covered with a streamlined waterproof shell to reduce the underwater resistance. Fig. \ref{Fig:Actuators} shows the five motors used on the second generation. The swing of legs is actuated by rotating the two DC motors (RE35) back and forth and using bevel gears transmission mechanism. The front rudder and two pectoral fins are controlled by three servo motors (Savox SC-1256TG) separately.

\subsection{Wheeled Locomotion}

\begin{figure}[thpb]
      \centering
      \includegraphics[width=70mm]{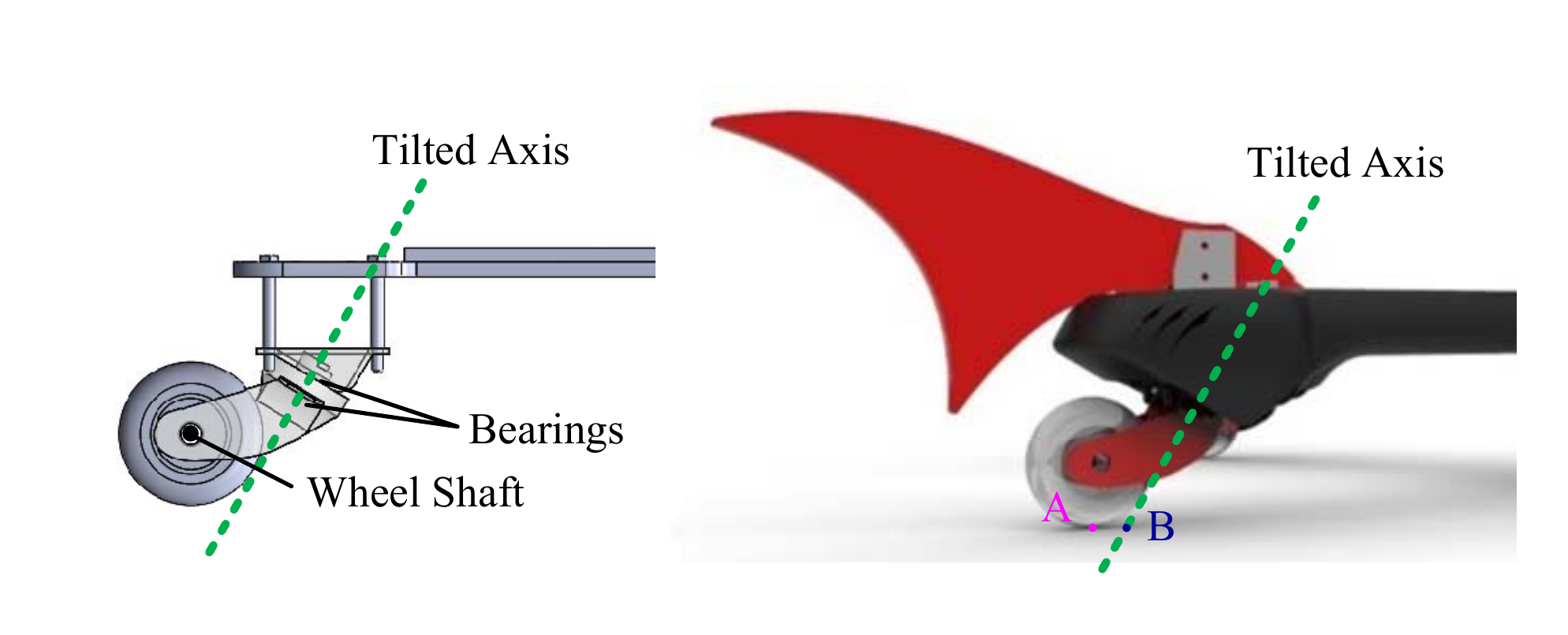}  
      \caption{The universal  wheel and caudal fin}
      \label{Fig:tail}
   \end{figure}

\begin{figure}[thpb]
      \centering
      \includegraphics[width=70mm]{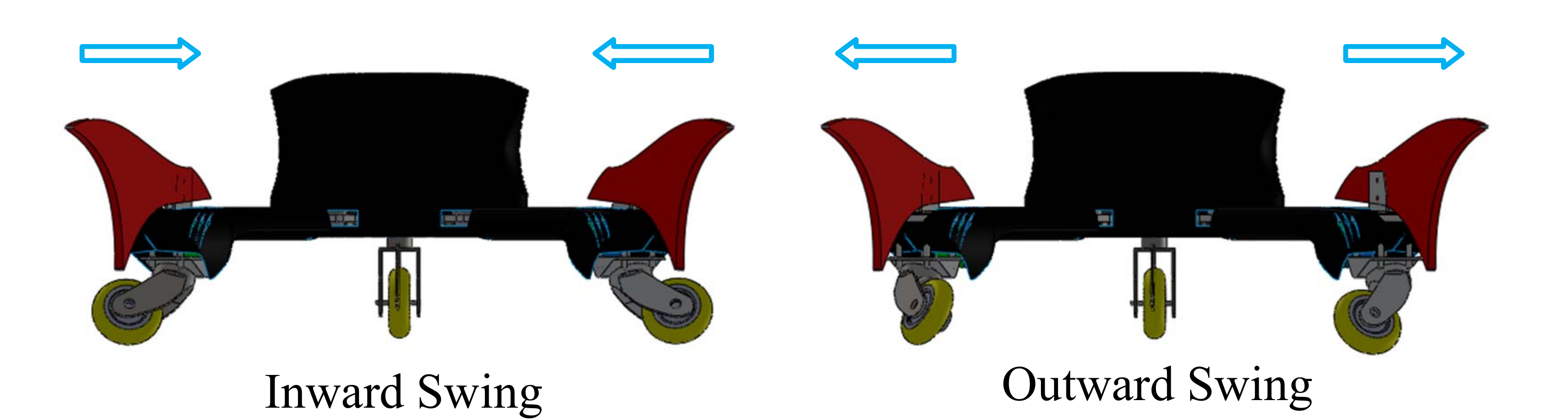}  
      \caption{Relationship between the anti-bias wheels and swing of legs}
      \label{Fig:Relationship}
   \end{figure}

As shown in Fig. \ref{Fig:tail}, the universal wheel connects to the bottom of the leg with bearings, so it can not only rotate around its own shaft, but also around the tilted axis of bearings integrally. The touch point A of the wheel with the ground is at the rear of the intersection B of the tilted axis with the ground. Therefore, the whole wheel will rotate around the tilted axis due to the friction between the wheel and the ground when the leg is swinging. Since the rotation direction of the whole wheel is different from the swing direction of the leg, this special universal wheel is called anti-bias wheel. In-toed gait and out-toed gait appear alternately during a swing period (see Fig .\ref{Fig:Relationship}). The propulsion comes from the friction between the wheel and  ground.

\subsection{Caudal Fin Undulation}

\begin{figure}[thpb]
      \centering
      \includegraphics[width=70mm]{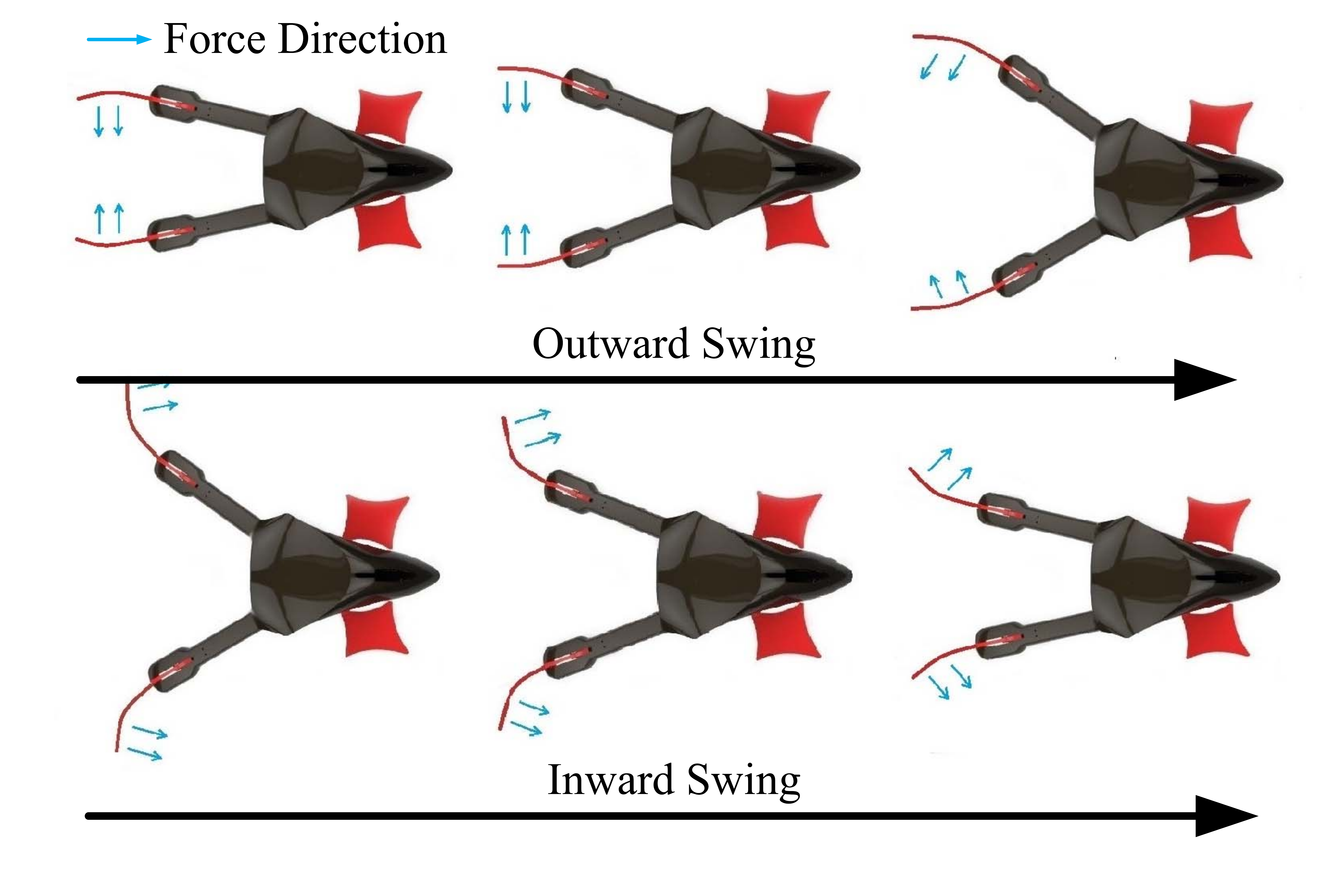}  
      \caption{Deformations of the caudal fins during a swing period}
      \label{Fig:Relationship force}
   \end{figure}

Most of BCF fish robots adopt serial open chain mechanisms, i.e. using linkages to fit the body curves. NAF-II, a two-DOF (one controlled DOF and one passive DOF) BCF robot, uses a rigid caudal fin to provide the propulsion \cite{c20}. A robust gait controller is used for steady swimming and direction control due to the complexity of underwater environment. Paired long undulating fins often appear in MPF type fish robots, and fin rays oscillate alternately with specific amplitude and frequency \cite{c21}. The symmetrical undulating motion makes the locomotion more steady. Too many fin rays, however, need to be controlled, bringing the complexity to the system. Due to the limited controlled DOF of FroBot's leg, a flexible caudal fin is used to fit the undulating curve. The deformation of the caudal fin comes from the interaction force between the caudal fin and water, as shown in Fig. \ref{Fig:Relationship force}. In addition, the symmetrical dual-swing-legs propulsion can benefit the heading stability.

\subsection{Pectoral Fin and Front Rudder}

Wing theory, one of the most compelling subjects of fluid mechanics, is usually utilized to generate lift force. According to hydromechanics, the angle between flow and wing is called attack angle, and the water active force is influenced by the size of the attack angle. Therefore, the attack angles of pectoral fins and front rudder can be controlled to achieve pitch, roll and yaw movement in water  \cite{c22}. As shown in Fig. \ref{Fig:Relationship angle}, the bilaterally equal attack angle can lead to pitch movement, and the bilaterally opposite attack angles can lead to distinct roll movement. When the servo motor of the front rudder rotates at a certain angle $\gamma$ on the ground or underwater, the drag force or resistance  $f_r$ can be decomposed into $f_y$ in the heading direction and $f_x$ in the lateral direction, as shown in Fig. \ref{Fig:Force analysis}. $f_y$ can be offset by propulsion, while $f_x$ provides steering moment, achieving the yaw movement of the robot.

\newcounter{TempEqCnt}                         
\setcounter{TempEqCnt}{\value{equation}} 
\setcounter{equation}{7}                           
\begin{figure*}[hb]
\begin{equation}\label{eq:the acceleration}
\begin{split}
\resizebox{0.95\hsize}{!}{$
	a_o= \left[\dfrac{(G_pL_G+2GL_1\cos\psi_2+2GL_2\cos\phi)(2\mu_{roll}-2\cos(\phi+\beta)\mu_{roll} +2\tan\lambda \sin\beta \sin(\phi+\beta) )}{(2L_1\cos\psi_2+2L_2\cos\phi-2\tan\lambda \sin\beta \sin(\phi+\beta) h)(G_p+2G) }
-\dfrac{K_{air}V_p}{(G_p+2G) }-\mu_{roll}\right]g
$}
\end{split}
\end{equation}
\end{figure*}
\setcounter{equation}{\value{TempEqCnt}} 

\begin{figure}[thpb]
      \centering
      \includegraphics[width=80mm]{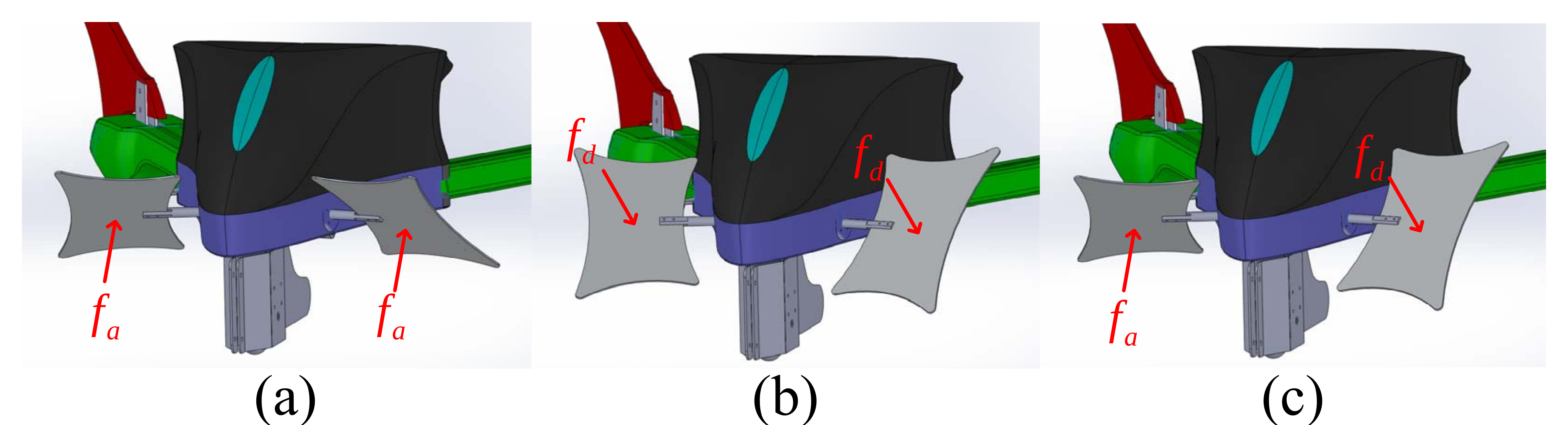}  
      \caption{Relationship between attack angles of pectoral fins and pitch((a) and (b)), roll(c) movement. Here  $f_a$ stands for lift force,  $f_d$  represents down force.}
      \label{Fig:Relationship angle}
   \end{figure}

\begin{figure}[thpb]
      \centering
      \includegraphics[width=50mm]{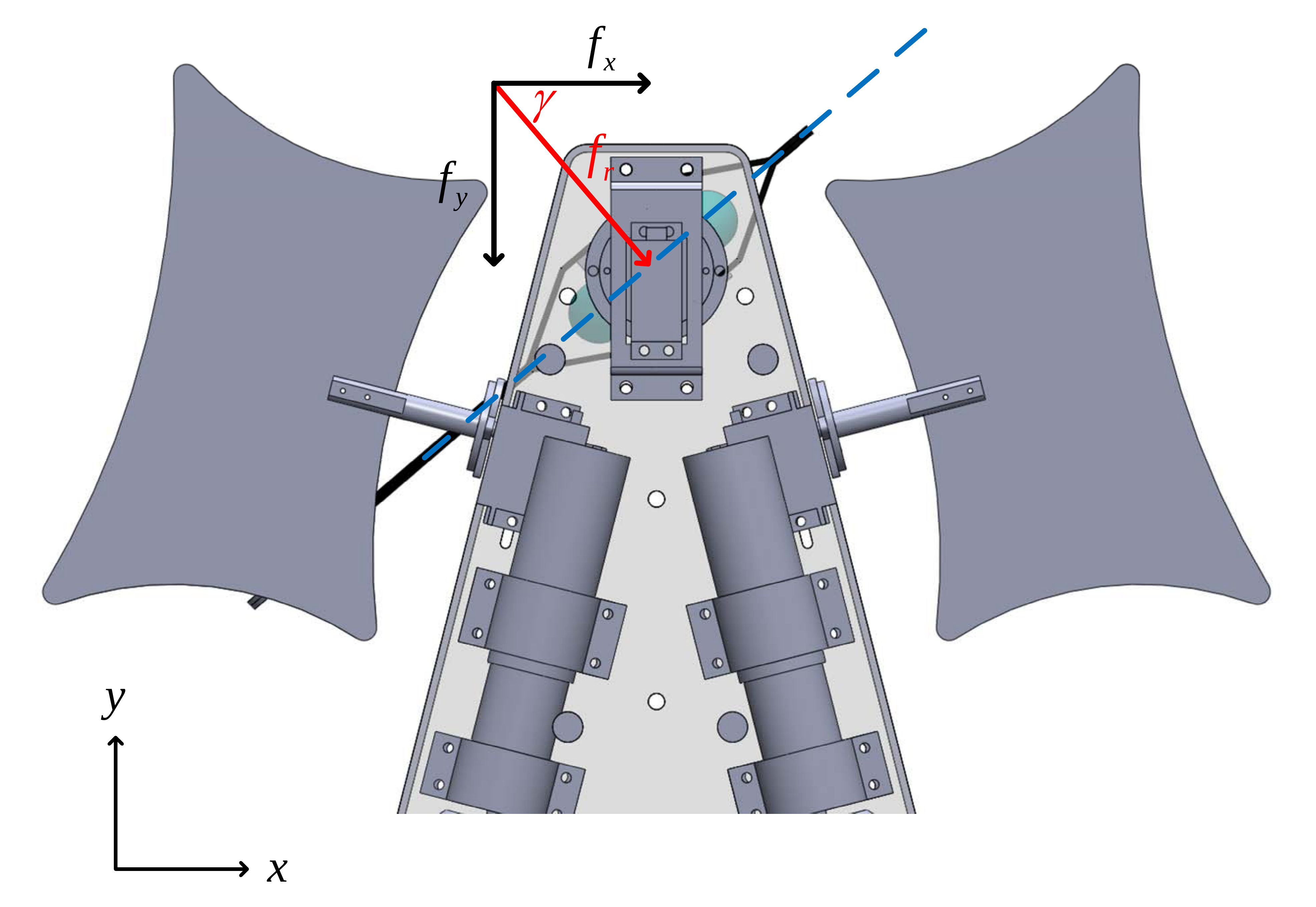}  
      \caption{Force analysis in yaw movement}
      \label{Fig:Force analysis}
   \end{figure}

\section{Dynamic Model on Land}

In this section, we focus on the dynamic model of FroBot on land. The propulsion for moving forward relies on the friction between the anti-bias wheels and ground.

\subsection{Force Analysis} 

\begin{figure}[thpb]
      \centering
      \includegraphics[width=80mm]{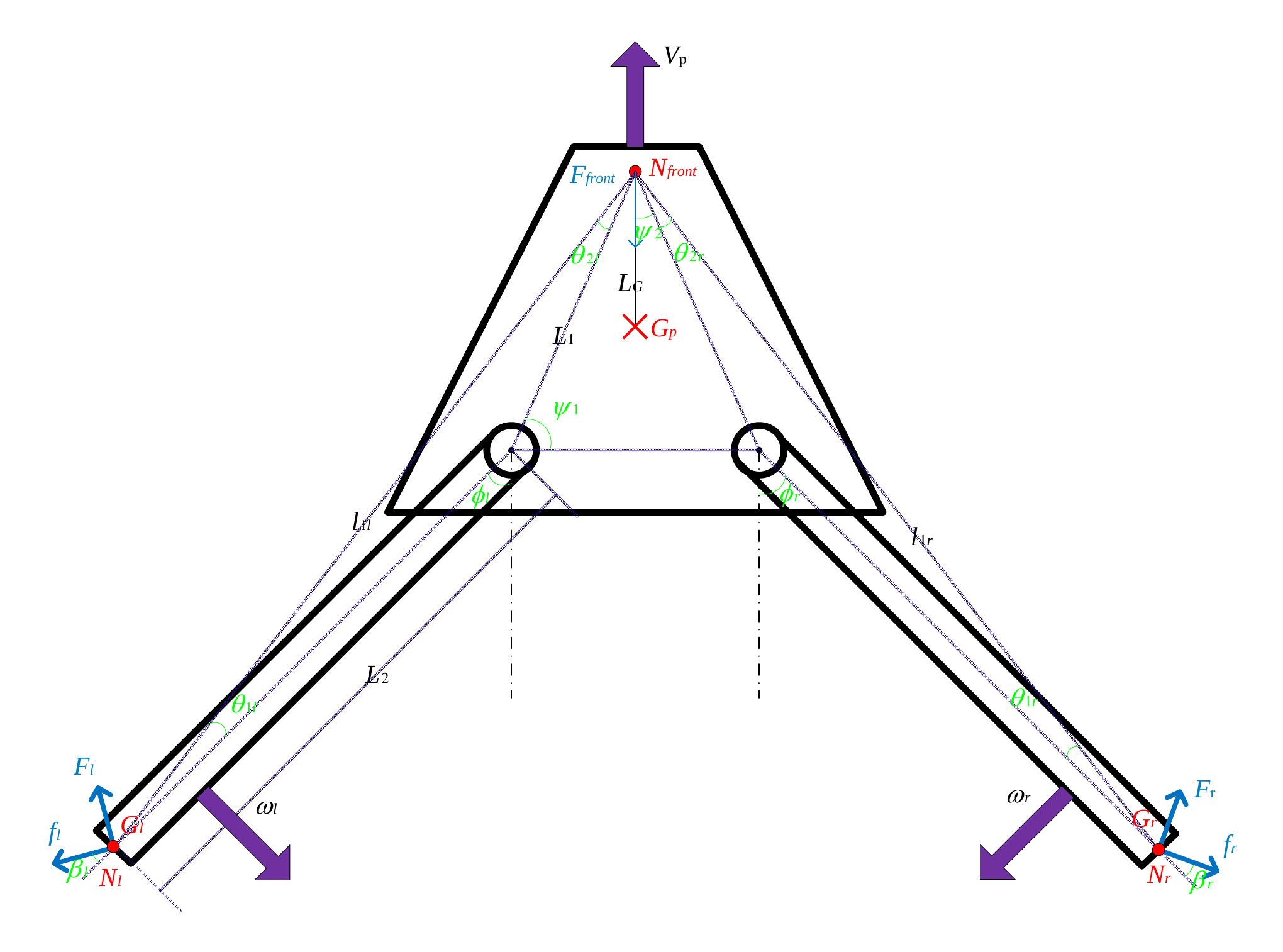}  
      \caption{Overall force diagram}
      \label{Fig11}
   \end{figure}

As shown in Fig. \ref{Fig11}, when the robot's legs are swinging, the forces mainly include:
\begin{itemize}
\item Tangential friction between the anti-bias wheels and ground.
\item Forces of vertical rolling friction between wheels and ground.
\item Air drag.
\item Gravities of the head and legs.
\item Normal forces of front wheel and the anti-bias wheels.
\end{itemize}

Tangential friction between the anti-bias wheels and ground is the main source of propulsion. To model the friction, a group of parameters as well as their values are listed in Table \ref{Table1}.

\begin{table}
\centering
\caption{physical parameters in the dynamic model}
\begin{tabular}{c|c|c}
\hline
Symbol& Value & Meaning\\
\hline
\hline
$N_{front}$&  &Normal force of front wheel \\
\hline
$N_{l},N_{r}$ &  &Normal forces of two legs \\
\hline 
$G_p$ & $44.1N$ & Gravity of the supporting platform\\
\hline
$G_{l}=G_{r}$ &$7.35N$  & Gravities of two legs\\
\hline
$F_{l},F_{r}$ & & Anti-bias forces of two wheels\\
\hline
$ f_{front}$&  &\tabincell{c}{ Force of rolling friction between\\ front wheel and ground} \\
\hline
$f_{l},f_{r}$&  &\tabincell{c}{Forces of rolling friction between\\ the anti-bias wheels and ground} \\  
\hline
$\omega_l,\omega_r $& & Angular velocity of two legs\\
\hline
$\beta_l,\beta_r$&  &Deflected angles of the anti-bias wheels\\
\hline
$\theta_{1l},\theta_{1r}$&  &Angle variables\\
\hline
$\phi_l,\phi_r$&  &\tabincell{c}{Angles between leg\\ and heading direction}  \\ 
\hline
$\psi_1$& $7\pi/{18}$ &Constant\\
\hline
$\psi_2$& ${\pi}/{9}$ &Constant\\
\hline
$L_1$& $0.25m$ &Constant \\
\hline
$L_2$ & $0.30m$ & Length of leg\\
\hline
$L_G$ & $0.12m$ &\tabincell{c}{ Length between front wheel\\ and barycenter of platform}  \\
 \hline
$l_l,l_r$   & &\tabincell{c}{ Length between front wheel\\ and anti-bias wheel}  \\ 
\hline
$h$ & $0.1m$ & Height of barycenter\\
\hline
$V_p$ &  &Velocity of supporting platform\\
\hline
\end{tabular}
\label{Table1} 
\end{table}

\subsection{Dynamic Model}

In order to simplify the dynamic model, following assumptions are given:
\begin{itemize}
\item  Acting points of $F_l; F_r; N_l; N_r; G_l $ and $ G_r$ all locate at the contact points of the anti-bias wheels with the ground.
\item The applied moment of $f_l$ and $f_r$ can be ignored due to the negligible effectiveness compared with $f_{front}$.
\item Deflected angles $\beta_l$ and $\beta_r$ change instantly during the reversing of legs.
\item Only the horizontal locomotion is taken into consideration.
\end{itemize}

According to the sine-cosine law, the force and moment balance equations can be established as:

\begin{equation}\label{eq:balance equations}
\begin{split}
l_lN_l&\cos(\psi_2+\theta_{2l})+l_rN_r\cos(\psi_2+\theta_{2r})=\\
         & G_pL_G+G_ll_l\cos(\psi_2+\theta_{2l})+G_rl_r\cos(\psi_2+\theta_{2r})\\
         &~~~~~+F_lh\sin(\phi_l+\beta_l)+F_rh\sin(\phi_r+\beta_r)
\end{split}
\end{equation}

In order to keep balance, the two legs have to swing symmetrically. Therefore, some variables can be unified as:
\begin{equation*}
\begin{split}
&N_l=N_r=N,~G_l=G_r=G,~\theta_{2l}=\theta_{2r}=\theta_2\\
&\theta_{1l}=\theta_{1r}=\theta_1,~l_l=l_r=l,~\omega_l=\omega_r=\omega\\
&\phi_l=\phi_r=\phi,~\beta_l=\beta_r=\beta,~{F_l} = {F_r} = F,~{f_l} = {f_r} = {f_{leg}}
\end{split}
\end{equation*}

As described in \cite{c23}:
\begin{equation}\label{eq:anti-bias force}
\begin{split}
F = N  \tan\lambda \sin\beta 
\end{split}
\end{equation}
where $\lambda$ is the angle between tilted axis and ground.
The anti-bias force $F$, whose direction is orthogonal to rolling direction, is the main component of the friction between the anti-bias wheels and ground.

Then the propulsion of the robot is :
\begin{equation}\label{eq:propulsion}
\begin{split}
\resizebox{0.9\hsize}{!}{$
{F_{forward}}=2F\sin(\phi+\beta)= 2N \tan\lambda  \sin\beta  \sin( {\phi  + \beta })
$}
\end{split}
\end{equation}

We make a definition here that $\beta$  is positive when the anti-bias wheel deflects inwards, while  negative  inversely. From \eqref{eq:balance equations} and \eqref{eq:propulsion}, we have:
\begin{equation}\label{eq:supportive}
\begin{split}
\resizebox{0.9\hsize}{!}{$ 
N = \frac{{{G_p}{L_G} + 2{\rm{G}}{L_1} {\rm{cos}}{{\rm{\psi }}_2} + 2{\rm{G}}{L_2}  {\rm{cos}}\phi }}{{2{L_1} {\rm{cos}}{{\rm{\psi }}_2} + 2{L_2} {\rm{cos}}\phi  - 2\tan\lambda  \sin\beta  \sin \left( {\phi  + \beta } \right)h}}
$}
\end{split}
\end{equation}

The force of rolling friction between wheels and ground  $f_{roll}$ and air drag  $f_{air}$ are the main drag forces, hence the total resistance is:

\begin{equation}\label{eq: resistance}
\begin{split}
f = {f_{roll}} + &{f_{air}}=\left( {Gp + 2G - 2N} \right){\mu _{roll}}\\
 &+ 2N{\mu _{roll}} \cos\left( {\phi  + \beta } \right){\rm{ + }}{K_{air}}{V_p}
\end{split}
\end{equation}
where $\mu_{roll}$ is the coefficient of rolling friction and has a value of 0.02, and $K_{air}$ is the coefficient of air resistance and has a value of 0.1.

Let  $a_o$ denote the acceleration of mass center, then according to the Newton's second law, we have:

\begin{equation}\label{eq:Newton’s second law}
\begin{split}
{F_{forward}} - f = \frac{{{G_p} + 2G}}{g}{a_o}
\end{split}
\end{equation}

From \eqref{eq: resistance} and \eqref{eq:Newton’s second law}, we get:

\begin{equation}\label{eq:newsupportive}
\begin{split}
\resizebox{0.9\hsize}{!}{$
N = \frac{{\frac{{{G_p} + 2G}}{g}{a_o} + ({G_p} + 2G){\mu _{roll}} + {K_{air}}{V_p}}}{{2{\mu _{roll}} - 2 \cos \left( {\phi  + \beta } \right){\mu _{roll}} + 2tan\lambda  \sin\beta  \sin \left( {\phi  + \beta } \right)}}$}
\end{split}
\end{equation}

According to \eqref{eq:supportive} and  \eqref{eq:newsupportive}, we can obtain \eqref{eq:the acceleration}. The equation reveals that the acceleration of the robot is decided by $\phi$  and $\beta$ . Note that $\beta$  is related to $\phi$   and $\omega$  , we can control the acceleration through changing $\omega$ , thus control the forward velocity.

 When the legs swing, the friction has two existence forms: static friction and sliding friction. If the friction manifests as sliding friction, we have:
\setcounter{equation}{8}
\begin{equation}\label{eq:friction}
\begin{split}
F  = N \tan\lambda  \sin\beta = N {\mu _{slide}}
\end{split}
\end{equation}
where $\mu_{slide}$ is the coefficient of sliding friction, then:
\begin{equation}\label{eq:coefficient of sliding friction}
\begin{split}
 \tan\lambda  \sin\beta  ={\mu _{slide}}
\end{split}
\end{equation}

As the experiments were conducted on the tile floor, we measured that ${\mu _{slide}} = 0.25$, and $\lambda  = {39\pi}/360 $, then we can get $\beta  ={\pi}/{4} $. As we know, the maximum static friction force is slightly greater than the force of sliding friction. Here we ignore the slight difference and let the maximum static friction force equal to the force of sliding friction, then  ${\beta _{max}} = {\pi}/{4} $.

\begin{figure}[thpb]
      \centering
      \includegraphics[width=85mm]{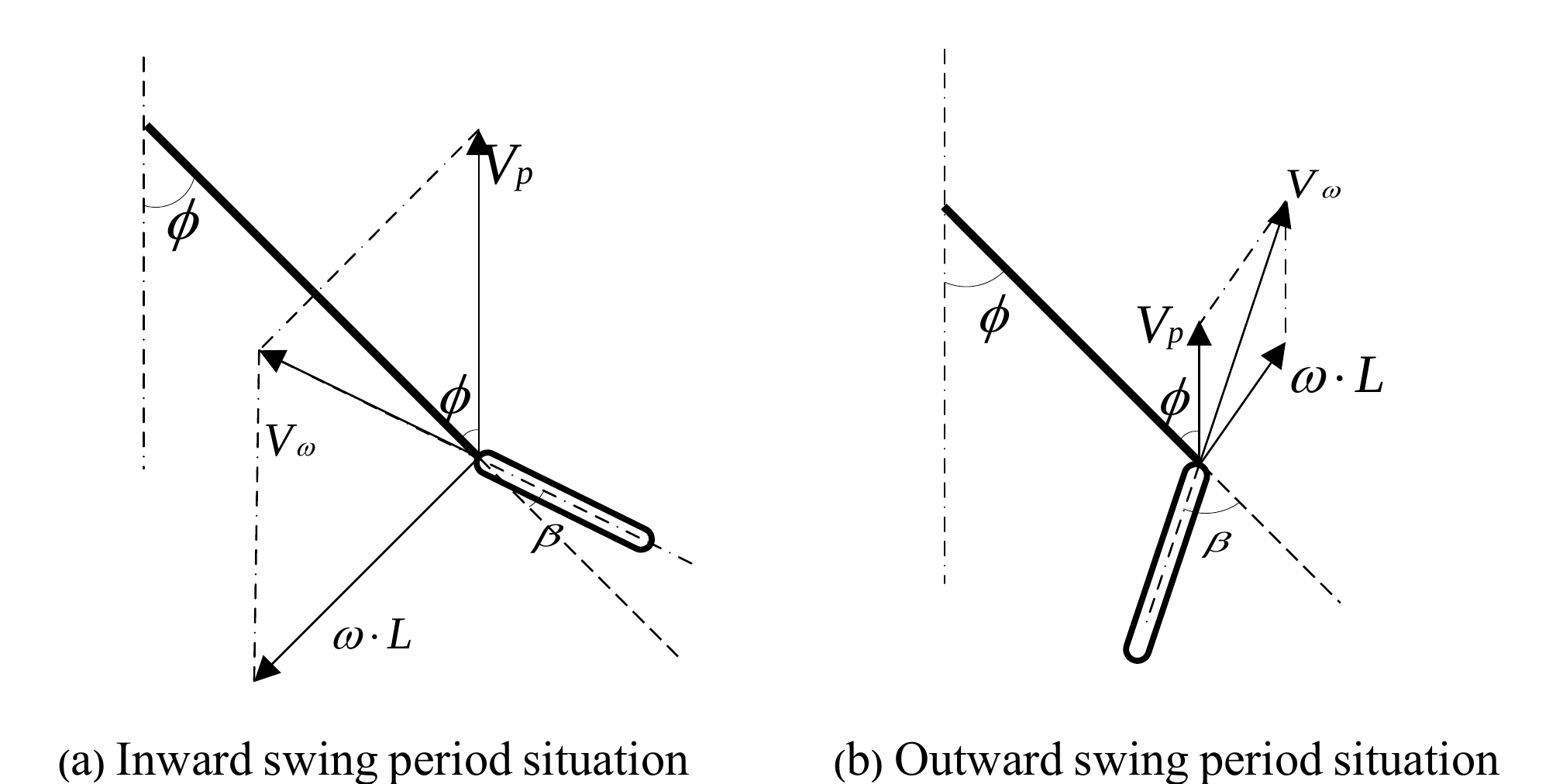}  
      \caption{Composition of velocity when the legs swing}
      \label{Fig12}
   \end{figure}

The composition of velocity when the legs swing is shown in Fig. \ref{Fig12}. Let  $V_p$ denote the velocity of the robot,   $\omega$ denote the angular velocity of legs and  $V_{\omega}$ denote the rolling velocity of the anti-bias wheels. The anti-bias wheels deflect outwards during inward swing period (see Fig. \ref{Fig12} (a)), while they deflect inwards during outward swing period (see Fig. \ref{Fig12} (b)). Analysis shows that  $\beta $  is highly related with   $V_p$  and  $\omega$, the relationship being:

\begin{equation}
\begin{split}
	\beta_{in~}=\begin{cases}%
			~~\dfrac{\pi}{4},~~~~\qquad \text{if } \omega L>  \sqrt 2 V_p\sin(\phi + \dfrac{\pi}{4})\\
		~~\arctan(\dfrac{\omega L-V_p\sin \phi}{V_p \cos \phi}),~~ \text{otherwise}
		\end{cases}
\end{split}
\end{equation}

\begin{equation}
\begin{split}
\beta_{out}=\begin{cases}%
			-\dfrac{\pi}{4},~~~~\qquad \text{if } \omega L >  \sqrt 2 V_p\cos(\phi + \dfrac{\pi}{4})\\
			-\arctan(\dfrac{\omega L+V_p\sin \phi}{V_p \cos \phi}), ~~\text{otherwise}
		\end{cases}
\end{split}
\end{equation}
where $\beta_{in}$  denotes the deflected angle of the  anti-bias wheels during inward swing period, and  $\beta_{out}$ denotes the deflected angle during outward swing period.

Following parameters will affect the locomotion of the robot:

\begin{itemize}
\item \emph{Range of  $\phi$}

It is hard for the robot to keep balance when $\phi$  is too small due to the mechanical structure feature. Furthermore, little propulsion will be obtained when $\phi$   is too big. We found that when we set $\phi< \pi/36$ or $\phi> \pi/3$, the robot become unbalanced. Hence, we  set  $\phi$  with a range of $[3\pi /20,11\pi /40]$ and the robot can move steadily. Then the swing amplitude $Amp = \pi /8$. Unless otherwise stated, the analyses below all adopt this swing amplitude.  

\item \emph{size of   $\omega$}

The inward swing angular velocity $\omega_{in}$  and the outward swing angular velocity $\omega_{out}$   are the dominating parameters  that affect the velocity of the robot.

\item \emph{Pause time}

As we know, when we swim breaststroke, we often stop for a period of time before a breaststroke kick. It is the same for the robot to keep balance and have a more smooth velocity. When we set the pause time as $150ms$ approximately, we can get a better smooth velocity and the robot can perform well.

\end{itemize}

\section{Experiments and Analysis on Land}

In order to evaluate the validity of the dynamic model established above,  we conducted some experiments to compare with the simulation results in this section.
\subsection{The Relationship Between  the Forward Velocity and Swing Angular Velocity}

\begin{figure}[thpb]
      \centering
      \includegraphics[width=85mm]{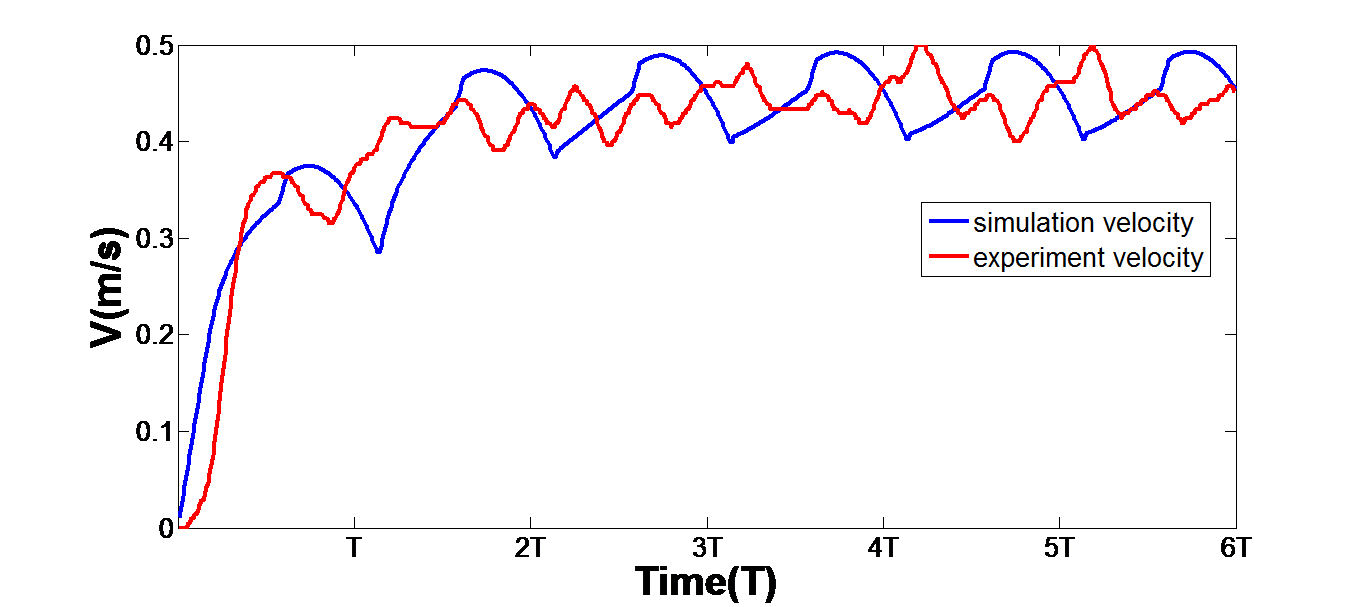}  
      \caption{Velocity response with given swing angular velocity}
      \label{Fig13}
   \end{figure}

Let ${\omega _{in}} = 1.17rad/s$ and ${\omega _{out}} = 0.975rad/s$, the simulation and experiment results of the robot  forward velocity are shown in Fig. \ref{Fig13}, where T is the swing period of legs. We can find that the velocity change trends of the two results are about the same, and the velocity tends to be stable after two swing periods.

\begin{figure}[thpb]
      \centering
      \includegraphics[width=85mm]{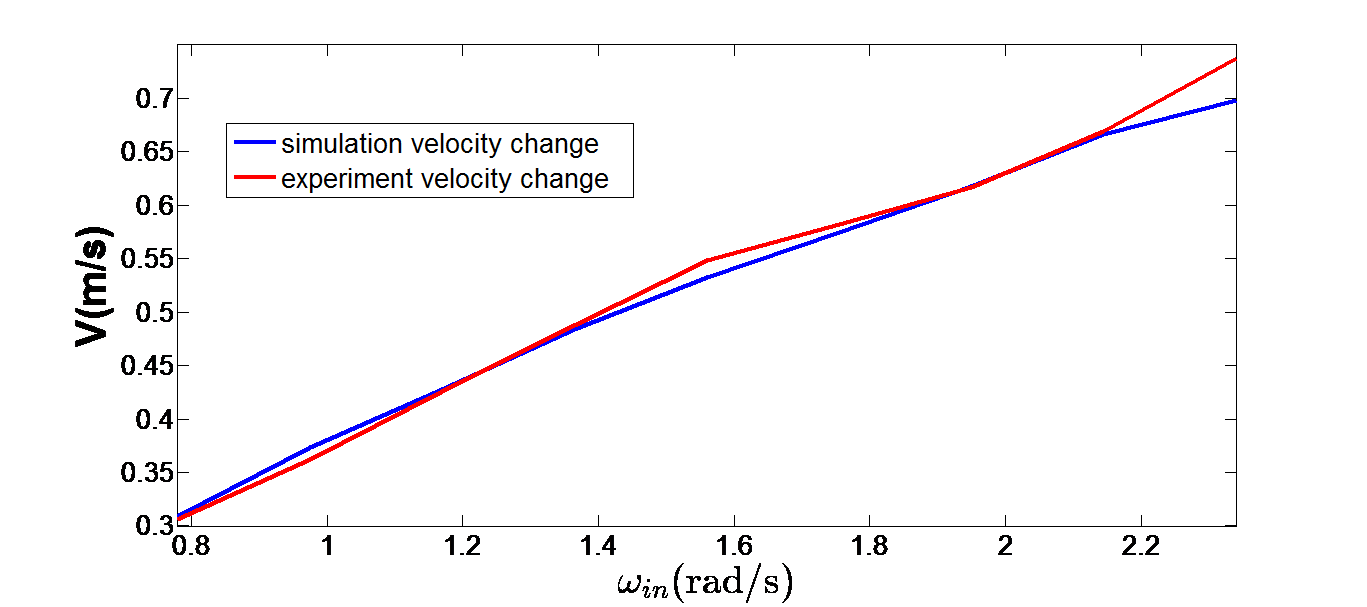}  
      \caption{Simulation and experiment results of different  $\omega_{in}$}
      \label{Fig14}
   \end{figure}

Let ${\omega _{out}} = 0.78rad/s$, ${\omega _{in}} $ change from  $0.78rad/s$ to  $2.34rad/s$,  we can obtain the steady forward velocity related to ${\omega _{in}}$, as shown in Fig. \ref{Fig14}.

\begin{figure}[thpb]
      \centering
      \includegraphics[width=85mm]{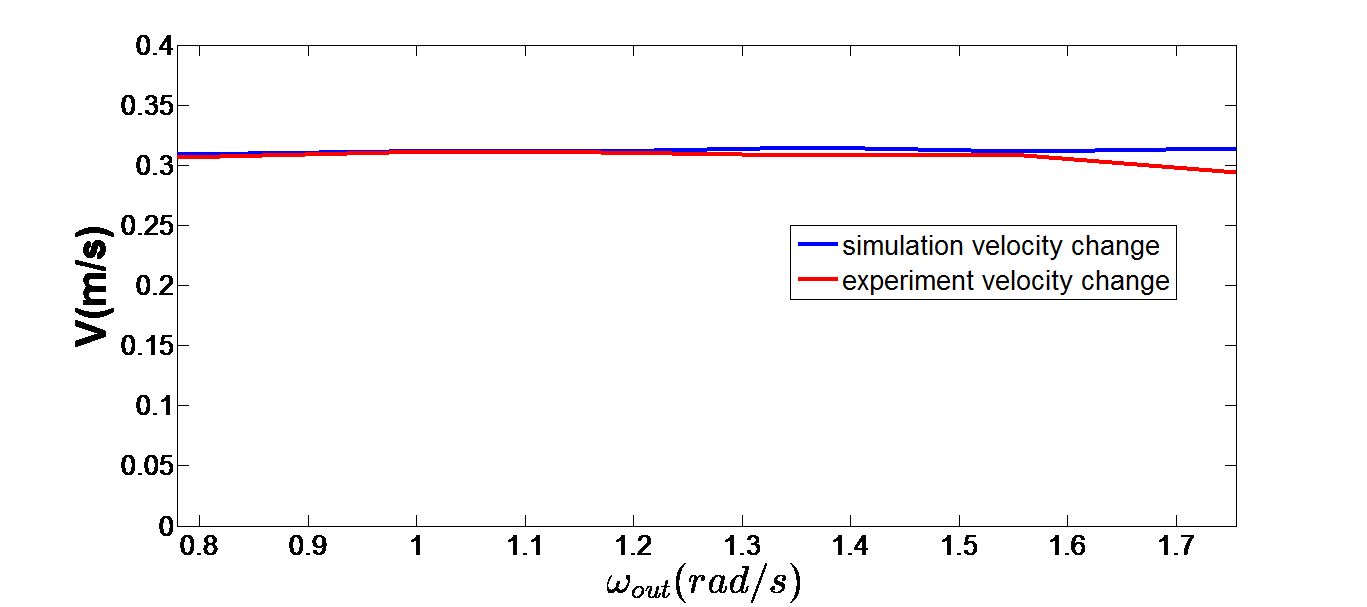}  
      \caption{Simulation and experiment results of different  $\omega_{out}$}
      \label{Fig15}
   \end{figure}

Let  ${\omega _{in}} = 0.78rad/s$, ${\omega _{out}} $ change from  $0.78rad/s$ to  $1.755rad/s$,  we can obtain the steady forward velocity related to ${\omega _{out}}$, as shown in Fig. \ref{Fig15}.

From simulation and experiment data, respectively, the robot forward velocity can be fitted as following experience formulas:

\begin{equation}\label{eq:varyspeed}
\begin{split}
RobotSpeed_{sim } &= 0.4017{\omega _{in}} - 0.01561{\omega _{in}}^2\\
& + 0.01339{\omega _{out}} + 0.004193{\omega _{out}}^2\\
RobotSpeed_{exp} &= {\rm{}}0.3645{\omega _{in}} - 0.01908{\omega _{in}}^2 \\
&+ 0.06387{\omega _{out}} - 0.016930{\omega _{out}}^2
\end{split}
\end{equation}
where $RobotSpeed$  is the forward velocity of the robot. In conclusion, the velocity is mainly dependent on  ${\omega _{in}}$ . However, too small ${\omega _{out}}$   will slow down the velocity of the robot severely and too big  ${\omega _{out}}$  will lead to the changing of static friction into sliding friction, and the legs may shake during the inward swing period, creating much challenge for control. Therefore, an applicable  ${\omega _{out}}$  is necessary. Unless otherwise specified, we always have ${\omega _{out}} = 0.78rad/s$  .

\subsection{Backward Locomotion and Climbing Up or Getting Down Slopes}

Backward locomotion is necessary for a mobile robot. As discussed above, the robot is able to move forward if the anti-bias wheels are installed normally. While if the wheels are fixed reversely, the directions of $\beta$  and friction  will be opposite. The dynamic model is similar to the forward locomotion. Simulation and experiment results have shown that FroBot is able to perform controllable backward locomotion. Thus we just need to attach the anti-bias wheels backwards to achieve the backward locomotion. In the future, we will use motors to reverse their direction, and then FroBot can do it  autonomously.

\begin{figure}[thpb]
      \centering
      \includegraphics[width=85mm]{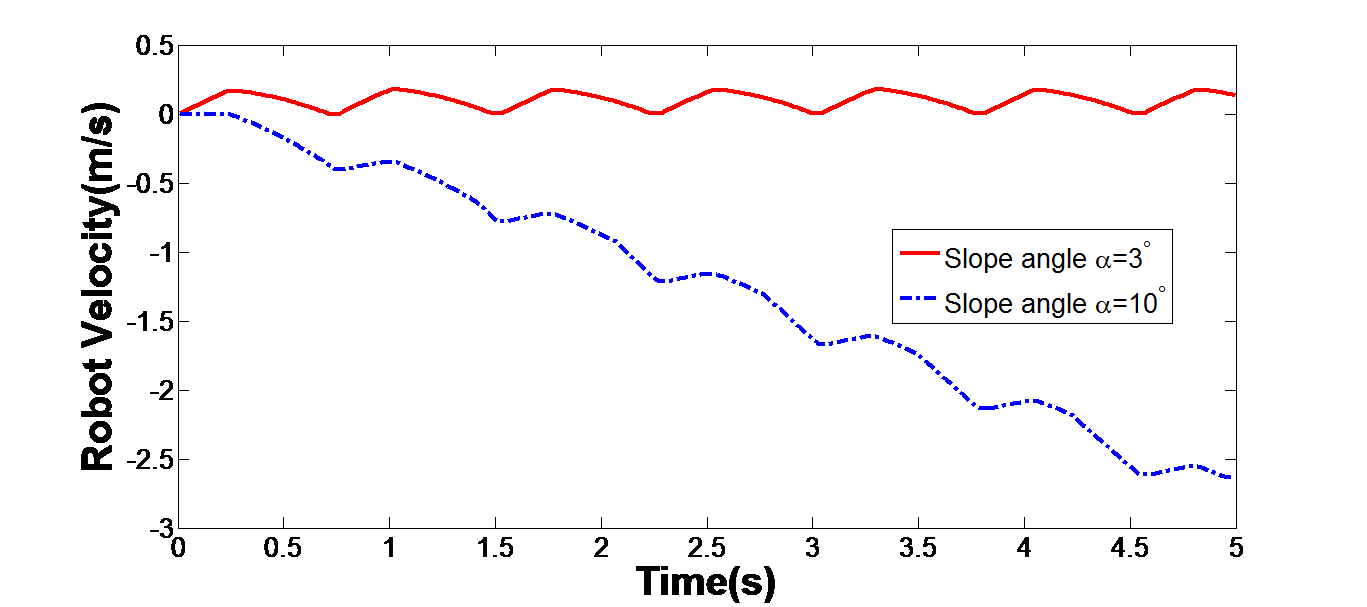}  
      \caption{Velocity responses with different slop angles}
      \label{Fig16}
   \end{figure}

Climbing up a slope is a challenge for FroBot due to the propulsion mechanism. With a slope of  $\alpha$ , all of its gravities should be taken into consideration. Simulation results show that the robot can't climb up the slope if  $\alpha \geqslant 4^\circ $  with ${\omega _{in}}$  taken as  $1.56rad/s$, as shown in Fig. \ref{Fig16}, and experiments give similar simulation results also.

\begin{figure}[thpb]
      \centering
      \includegraphics[width=85mm]{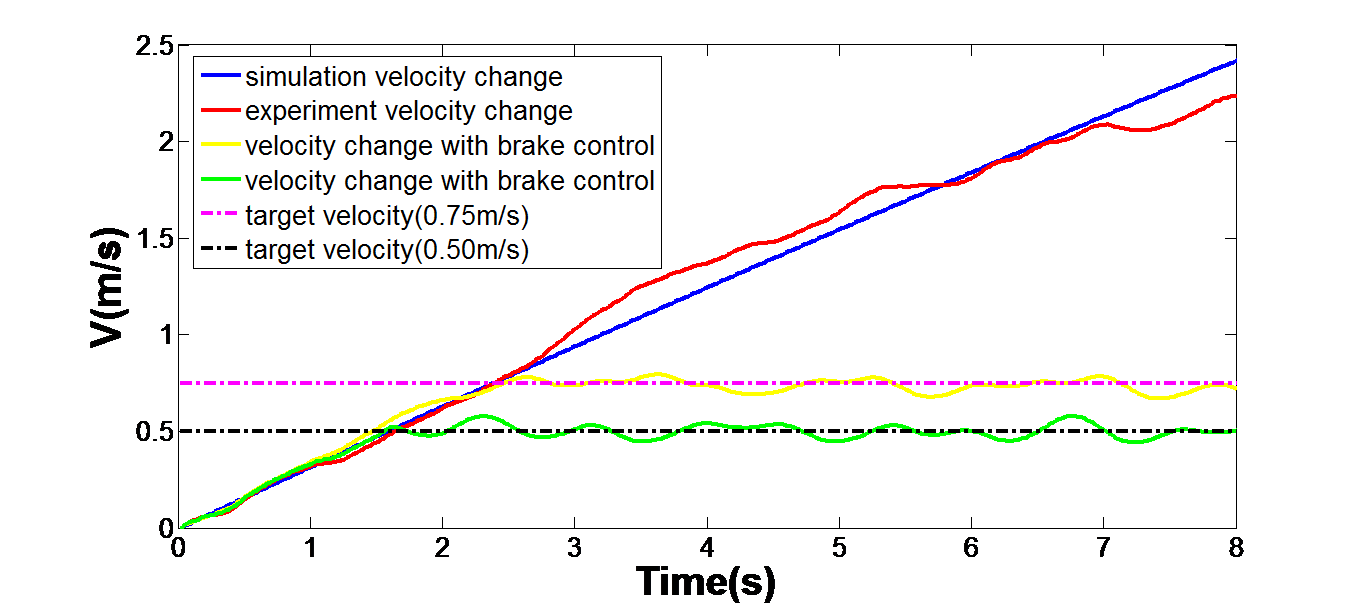}  
      \caption{Velocity responses with or without control when getting down slope}
      \label{Fig17}
   \end{figure}

The anti-bias wheel is a kind of universal wheel which can't be controlled to stop rolling when the robot is getting down a long slope. As we can imagine, the velocity will increase quickly if we don't utilize a brake apparatus. We chose a $9m$  long slope to conduct experiment, and kept $\phi {\rm{ = }}\pi {\rm{/5}}$ unchanged in the experiment. The results are shown in Fig. \ref{Fig17}.

\subsection{Smooth Velocity Control}
\begin{figure}[thpb]
      \centering
      \includegraphics[width=90mm]{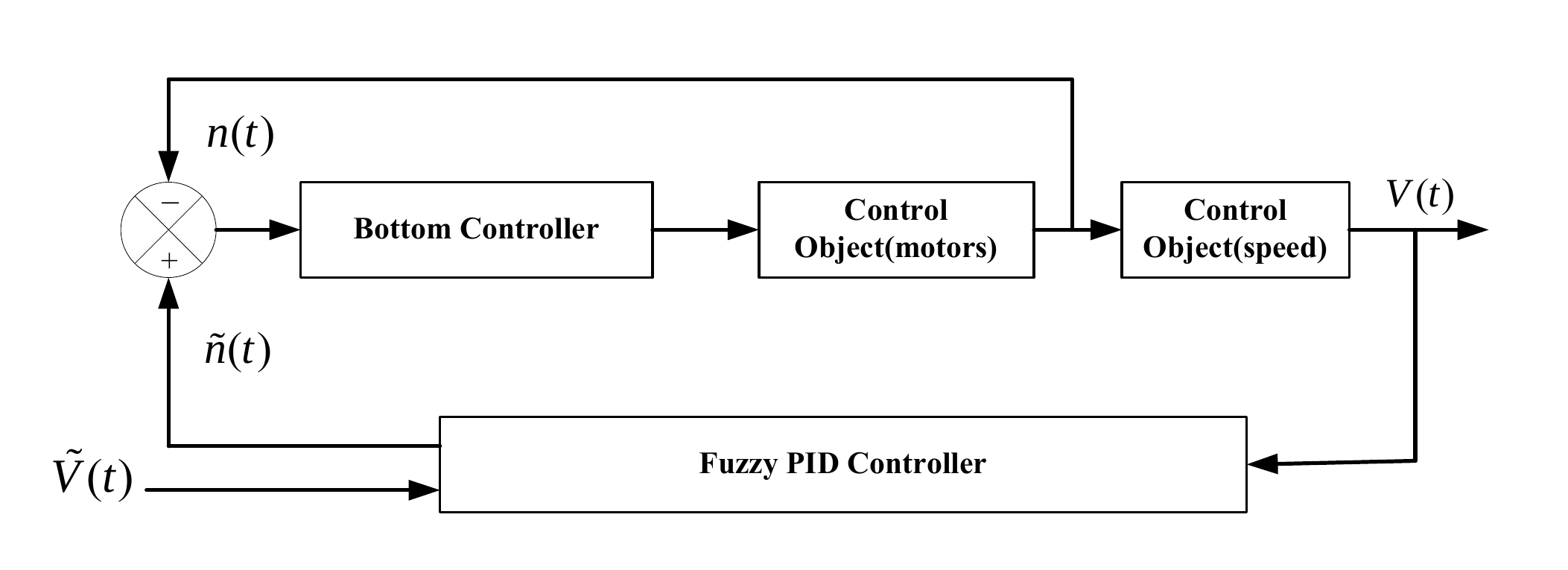}  
      \caption{Structure of control system}
      \label{Fig18}
   \end{figure}

As discussed above, the robot is a complicated, time-varying and nonlinear system, and the velocity fluctuates largely during the locomotion. Conventional PID controller is not able to yield satisfactory stable velocity control. Hence we chose a fuzzy PID controller \cite{c24}  to get a smooth velocity. As shown in Fig. \ref{Fig18}, the system consists of two closed loops. The inner loop adopts PID controller to control the motors, and the outer loop adopts the fuzzy PID controller to realize the velocity control.

\begin{figure}[thpb]
      \centering
      \includegraphics[width=85mm]{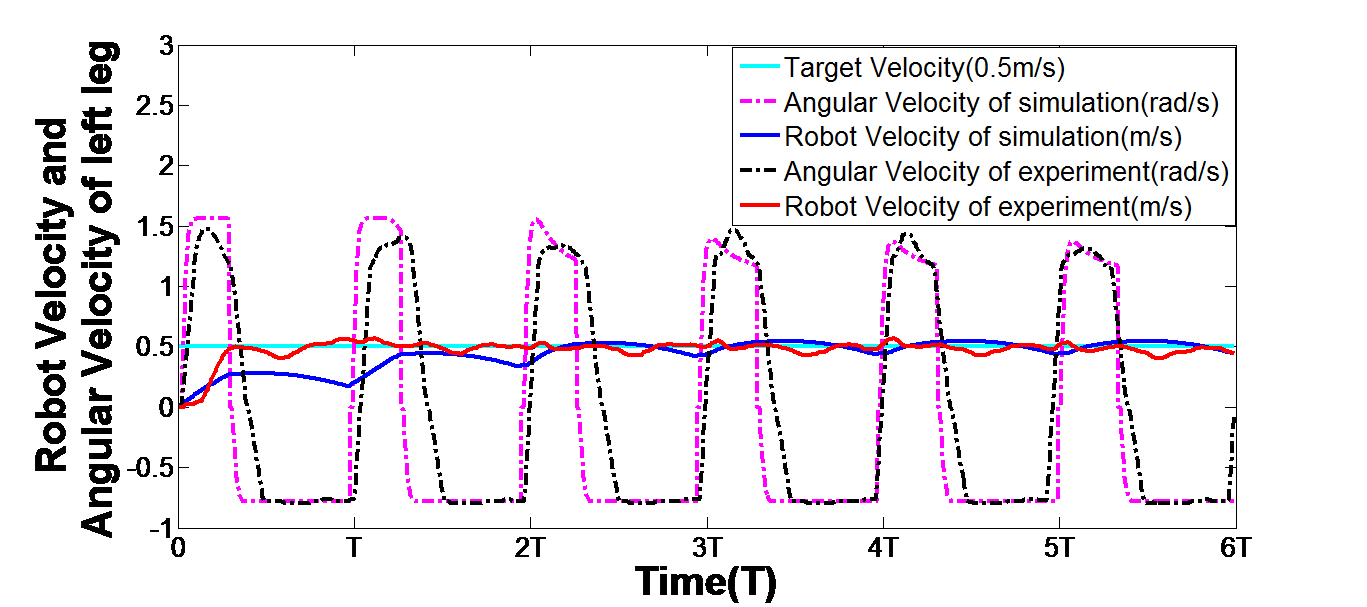}  
      \caption{Simulation and experiment results of fuzzy PID controller}
      \label{Fig19}
   \end{figure}

Given a target velocity of $0.5m/s$ ,  the results of the fuzzy PID controller are shown in Fig. \ref{Fig19}. The robot reaches the target velocity within a swing period, and  the actual velocity of the robot is able to track the target velocity.  

\begin{figure}[thpb]
      \centering
      \includegraphics[width=85mm]{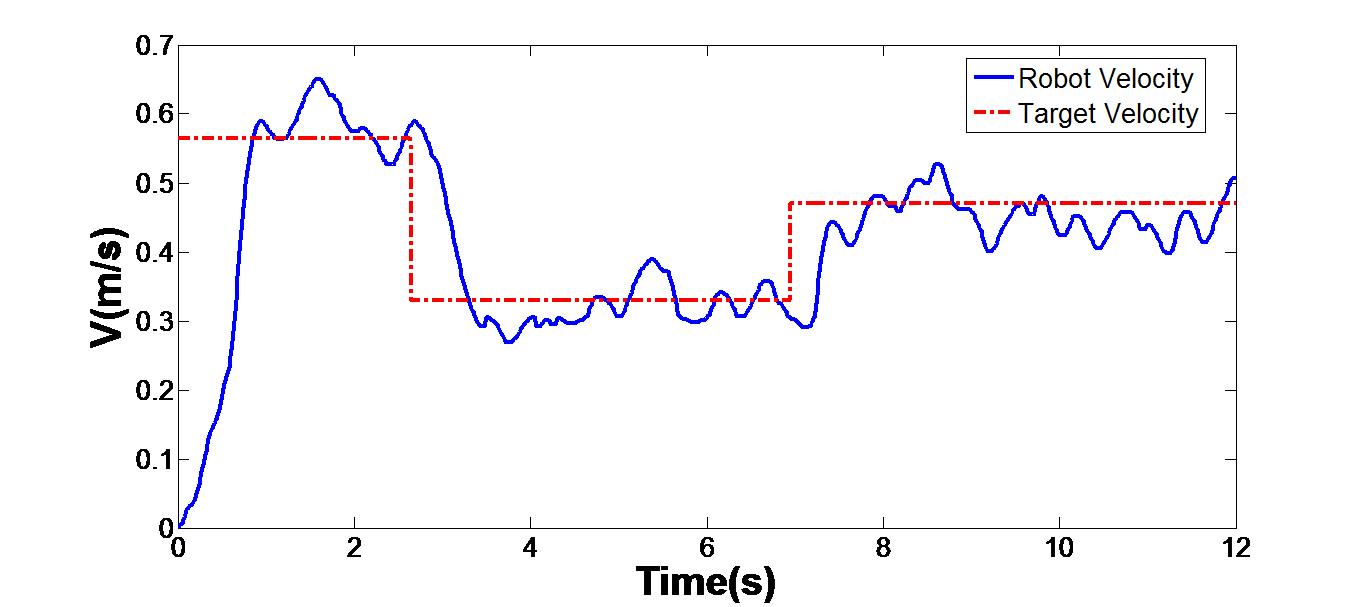}  
      \caption{Controller performance for step changes of target velocity}
      \label{Fig20}
   \end{figure}

Fig. \ref{Fig20} illustrates the controller performance for step changes of target velocity, and after each step the target velocity is held for a period of time to observe the step response and steady state behavior.

\begin{figure}[thpb]
      \centering
      \includegraphics[width=85mm]{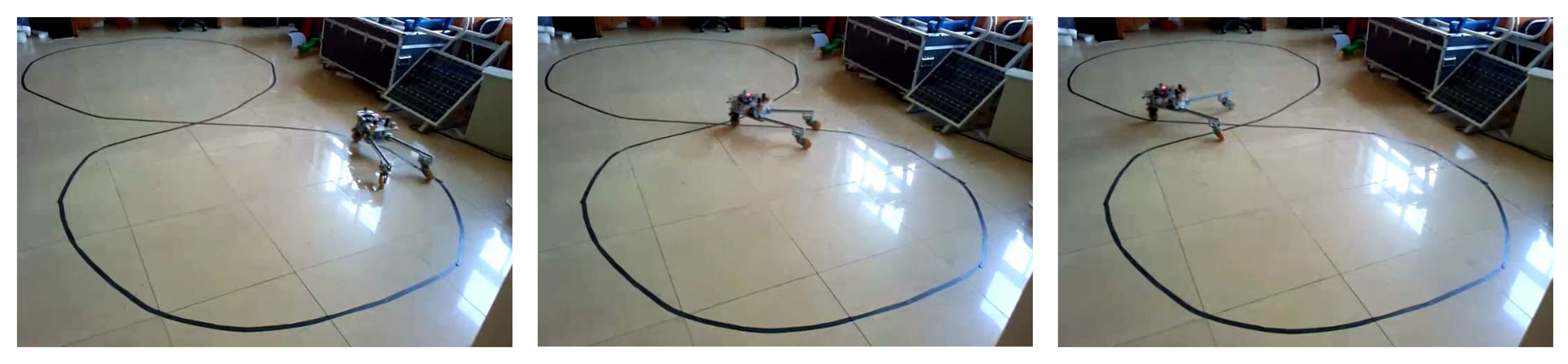}  
      \caption{Snapshots of trajectory following}
      \label{Fig23}
   \end{figure}
In order to verify the effect of motion control, we have performed some trajectory following experiments. A 8-shaped trajectory was presented (Fig. \ref{Fig23}), and a linear array CCD sensor was utilized to capture the relative position of trajectory. Forward direction is controlled by the front rudder, and closed loop of velocity is realized by the fuzzy PID controller. In the experiment, we set the target velocity of $0.47m/s$. The result shows that the velocity is controlled within $0.40m/s-0.55m/s$, and the average velocity is about $0.47m/s$.

\begin{figure}[thpb]
      \centering
      \includegraphics[width=85mm]{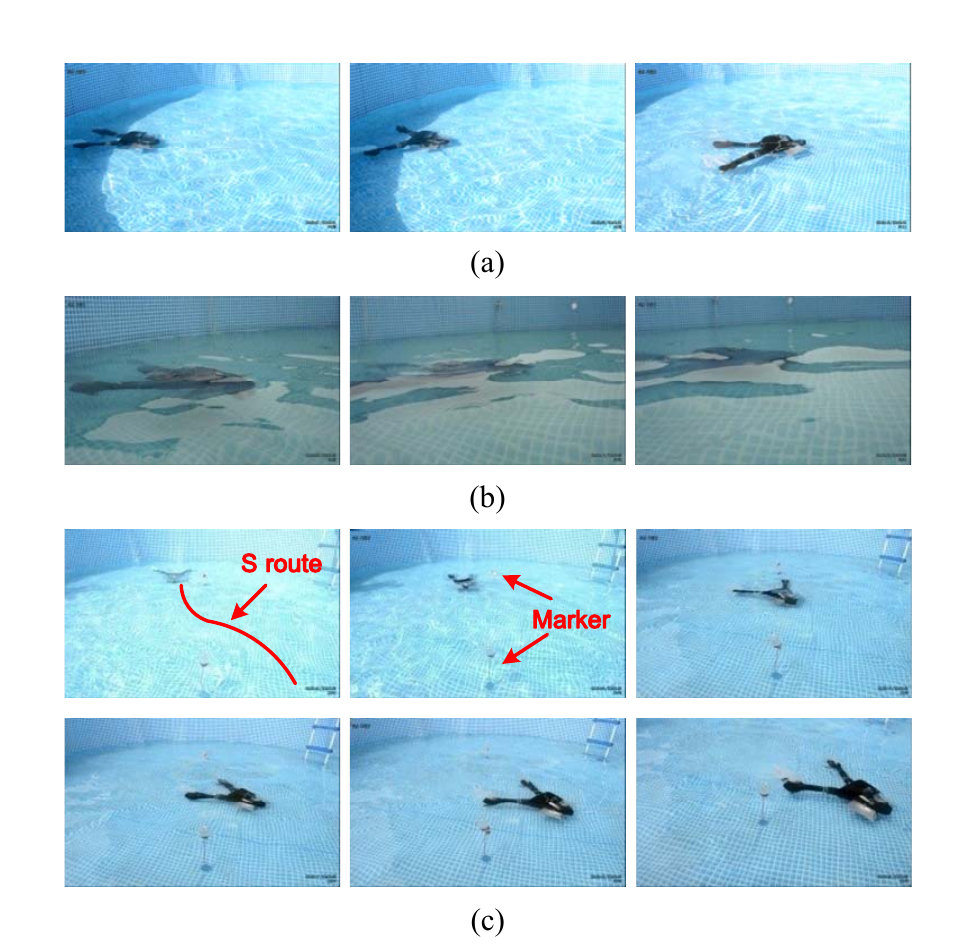}  
      \caption{Snapshots of underwater locomotion. (a) swimming straight, (b) descending, (c) swimming in S route.}
      \label{Fig24}
   \end{figure}

\subsection{Partial Features and Parameters of  FroBot}
The robot is approximately $70 cm$ long, $26 cm$ high, and  $23cm$ wide of head. Onboard battery would provide more than
two hours of continuous operation. The maximal velocity can reach $1m/s$ on land, and the turning radius is about $0.5m$ .

\section{Underwater Locomotion Experiment}

FroBot uses the same dual-swing-legs mechanism to propel itself in water. As the propulsion in water comes from the caudal fins, the above dynamic model established on land is not suitable for the aquatic environment. The robot has a waterproof shell made of ABS plastic and displaces water of $13kg$ approximately. Given the complexity of underwater environment and limited onboard sensors, we utilized remote control to conduct underwater experiment with terrestrial control methods. The experiment was conducted in a rounded pool with $5.5m$ in diameter and $1.0m$ in water depth. In the experiment, we observed that the flexible caudal fin was able to produce desired deformation to fit the expected curve. The results showed that the robot can implement forward swimming, yaw, pitch and roll movement, as shown in Fig.  \ref{Fig24}. Offline analysis of the experiment videos showed that the robot can reach a stable speed of $0.4m/s$.

\section{Conclusions and Future Work}

In this paper, we have proposed a frog-inspired amphibious robot. The propulsion comes from the anti-bias wheels on land and caudal fins in aquatic environment, and a pair of pectoral fins and a front rudder are used for attitude control. The key parameters that affect its locomotion on land are swing amplitude and frequency. Its dynamic model of forward locomotion, backward locomotion and climbing up or getting down a slope on land are analysed and verified. In addition, a fuzzy PID controller is adopted for a smooth speed control and trajectory following. To prove the rationality of the mechanical design and availability of control method, a series of underwater experiments are also conducted. 

\begin{figure}[thpb]
	\centering
	\includegraphics[width=85mm]{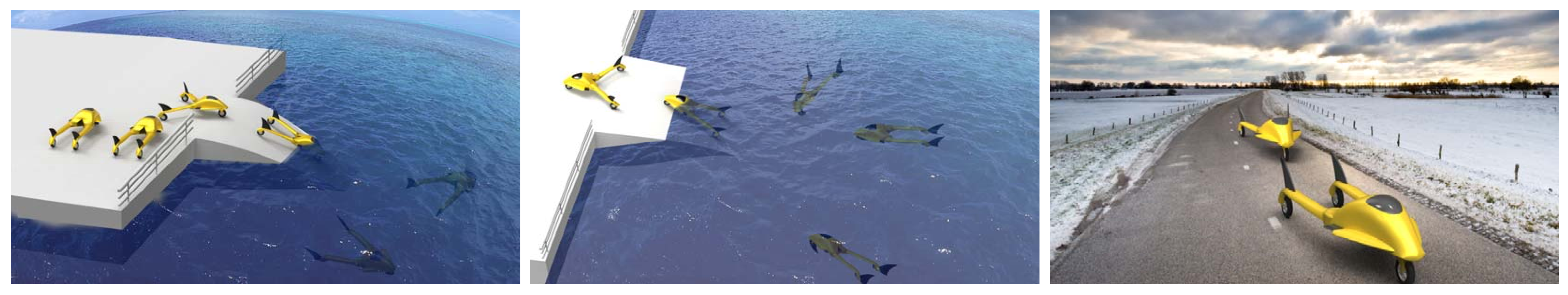}  
	\caption{Application prospect of FroBot}
	\label{improved}
\end{figure}

As FroBot adopts a new wheeled locomotion, making traversal and moving on soft surfaces difficult,  it has many limitations on land. Particularly, climbing up a slope exceeding 4 degrees is rather difficult due to the passive wheel structure. Our future work will focus on improving the mechanical structure. The momentum might be improved by adding a driving wheel or changing the wheel structure. Furthermore, rotating of the motors back and forth poses a challenge, so a linkage mechanism might be used in the next generation to avoid the reciprocating motion of motors. Backward locomotion needs to change the installation direction of the anti-bias wheels, and FroBot has limited maneuverability to produce more propulsion when it is in an outward swing period, so an actuator on each leg might be added to improve the performance of aquatic locomotion and realize the backward locomotion automatically. Frobot will have a  better application prospect with its limitations to be eliminated (see Fig. \ref{improved}).

\section{Acknowledgment}
The authors would like to thank Bai Kexian, Wang Meiling, Ma Hongbin, Lei Yong, Wang Zhili, Qiu Fan, Xie Shanshan, Li Donghai and all other members of BIT ININ Lab for their contribution to this work.

\end{document}